\documentclass[conference]{IEEEtran}
\IEEEoverridecommandlockouts
\usepackage{cite}
\usepackage[numbers,sort&compress]{natbib}

\usepackage[utf8]{inputenc}
\usepackage{amsmath,amssymb}
\usepackage{graphicx}
\usepackage{bm}
\usepackage{enumerate}
\usepackage{comment}
\usepackage{url}
\usepackage{color,soul}
\usepackage{subcaption}
\usepackage{float}

\definecolor{light-gray}{gray}{0.8}

\usepackage{amsmath,amssymb,amsfonts}
\usepackage{algorithmic}
\usepackage{graphicx}
\usepackage{textcomp}
\usepackage{subcaption}

\usepackage{colortbl}
\usepackage{multirow}
\usepackage{array}
\usepackage{booktabs}
\usepackage[table]{xcolor} 

\def\BibTeX{{\rm B\kern-.05em{\sc i\kern-.025em b}\kern-.08em
    T\kern-.1667em\lower.7ex\hbox{E}\kern-.125emX}}

\makeatletter
\newcommand{\linebreakand}{%
  \end{@IEEEauthorhalign}
  \hfill\mbox{}\par
  \mbox{}\hfill\begin{@IEEEauthorhalign}
}
\makeatother

\begin{document}

\title{Evaluating Modern Approaches in 3D Scene Reconstruction: NeRF vs Gaussian-Based Methods \\}

\author{

\small 

\begin{tabular}[t]{c@{\extracolsep{8em}}c} 

1\textsuperscript{st} Yiming Zhou  & 2\textsuperscript{nd} Zixuan Zeng \\
\textit{Saarland University of Applied Science} & \textit{Guangxi University} \\
Saarland, Germany & Nanning, China \\
yiming.zhou@htwsaar.de & axuan11722@foxmail.com \\

\\

3\textsuperscript{rd} Andi Chen & 4\textsuperscript{th} Xiaofan Zhou \\
\textit{Independent Researcher} & \textit{University of Florida} \\
Beijing, China & Gainesville, USA \\
real.andichen@gmail.com & zhouxiaofan2011@gmail.com \\

\\

5\textsuperscript{th} Haowei Ni & 6\textsuperscript{th} Shiyao Zhang \\
\textit{Columbia University} & \textit{Cornell University} \\
New York, USA & Ithaca, USA \\
hn2339@caa.columbia.edu & sz566@cornell.edu \\

\\

7\textsuperscript{th} Panfeng Li & 8\textsuperscript{th} Liangxi Liu \\
\textit{University of Michigan} & \textit{Northeastern University} \\
Ann Arbor, USA & Boston, USA \\
pfli@umich.edu & liu.liangx@northeastern.edu \\

\\

9\textsuperscript{th} Mengyao Zheng & 10\textsuperscript{th} Xupeng Chen \\
\textit{Harvard University} & \textit{New York University} \\
Boston, USA & New York, USA \\
mengyaozheng@alumni.harvard.edu & xc1490@nyu.edu \\

\end{tabular}
}

\maketitle

\begin{abstract}
Exploring the capabilities of Neural Radiance Fields (NeRF) and Gaussian-based methods in the context of 3D scene reconstruction, this study contrasts these modern approaches with traditional Simultaneous Localization and Mapping (SLAM) systems. Utilizing datasets such as Replica and ScanNet, we assess performance based on tracking accuracy, mapping fidelity, and view synthesis. Findings reveal that NeRF excels in view synthesis, offering unique capabilities in generating new perspectives from existing data, albeit at slower processing speeds. Conversely, Gaussian-based methods provide rapid processing and significant expressiveness but lack comprehensive scene completion. Enhanced by global optimization and loop closure techniques, newer methods like NICE-SLAM and SplaTAM not only surpass older frameworks such as ORB-SLAM2 in terms of robustness but also demonstrate superior performance in dynamic and complex environments. This comparative analysis bridges theoretical research with practical implications, shedding light on future developments in robust 3D scene reconstruction across various real-world applications.


\end{abstract}


\section{Introduction}
Recent advancements in Simultaneous Localization and Mapping (SLAM) have significantly leveraged deep learning to enhance the robustness and accuracy of 3D reconstruction and camera tracking. The use of neural implicit representations, as demonstrated by NICE-SLAM~\cite{zhu2022niceslam}, allows for detailed reconstructions of large-scale indoor scenes through a hierarchical grid-based encoding that efficiently manages local updates and optimizes scene representations within viewing frustums~\cite{peng2024xslam,zhu2022niceslam,zhang2023complex}. Additionally, the comprehensive evaluation in \cite{zhang2024deepgi} shows that the proposed method outperforms other state-of-the-art approaches, providing a novel solution for automated GI tract segmentation through the integration of specialized architectures.


Another critical advancement in SLAM is the introduction of global optimization techniques. GO-SLAM~\cite{zhang2023goslamg}, for example, integrates loop closure and full bundle adjustment in real-time, addressing limitations of previous works that perform these optimizations post-process~\cite{zhang2023goslamg, li2024ddnslam}. The continual integration of deep learning into SLAM processes not only enhances the adaptability of these systems but also significantly reduces the latency in processing complex datasets.


Despite these innovations, 3D reconstruction remains a complex challenge, with existing algorithms often excelling in one aspect while compromising in others. This paper focuses on comparing two state-of-the-art algorithms in the context of 3D reconstruction. We analyze Neural Radiance Fields (NeRF) and Gaussian-based methods, highlighting their respective advantages and disadvantages. NeRF, though slower, enables novel view synthesis, offering unprecedented capabilities in generating new perspectives from existing data. In contrast, Gaussian-based methods are noted for their speed and expressiveness but fall short in tasks like scene completion, which are crucial for comprehensive environmental understanding.

Comparative analyses between these innovative methods and traditional SLAM systems reveal significant performance improvements. Techniques like GO-SLAM~\cite{zhang2023goslamg} not only outperform older systems like ORB-SLAM2 and DROID-SLAM in tracking accuracy but also in the fidelity of 3D reconstructions \cite{zhang2023goslamg}, especially under challenging conditions that typically result in error accumulation and inconsistencies in traditional methods. Ultimately, this study aims to bridge the gap between theoretical advancements and practical applications, offering insights that could steer future developments in the field of 3D scene reconstruction.

\section{Methods}
\subsection{Data Collection}
\subsubsection{Replica \cite{straub2019replica}} 
The Replica dataset features 18 photorealistic 3D indoor scenes, encompassing a diverse array of environments such as offices, hotel rooms, and apartments. These scenes are richly detailed, showcasing dense meshes, high dynamic range (HDR) textures, semantic layers, and reflective properties. Created using a custom-built RGB-D capture rig that synchronizes IMU, RGB, and IR data, this dataset excels in providing comprehensive spatial and textural data, ideal for testing the nuanced rendering capabilities of NeRF-based models.

\subsubsection{ScanNet \cite{dai2017scannet}}
ScanNet encompasses over 2.5 million views from 1,513 3D scans of varied indoor locations, equipped with detailed annotations. The dataset uses structural sensors connected to handheld devices such as iPads, capturing extensive environmental data. The offline processing phase of ScanNet enhances these captures into comprehensive 3D scene reconstructions that include precise 6-DoF camera poses and semantic labeling. 


\subsection{Model Training}
\subsubsection{NeRF Based}
\paragraph{NICE-SLAM}
NICE-SLAM~\cite{zhu2022niceslam} employs a structured approach to 3D scene reconstruction, leveraging multi-level voxel grids to enhance detail capture and scalability. This hierarchical grid structure, as illustrated in Fig. 1, is pivotal in addressing common issues such as the over-smoothing of scene features. By updating only the grid features that are visible, NICE-SLAM significantly improves optimization precision and operational efficiency, contrasting sharply with methods like iMAP which rely on global updates and may suffer from inefficiency and error propagation. The algorithm typically utilizes a set of predefined parameters for constructing the voxel grids \cite{10229827}. These include setting the grid resolutions at different hierarchy levels to balance detail and computational efficiency. For instance, coarse grids capture basic structural outlines, while finer grids focus on detailed textures and objects. 


However, NICE-SLAM is not without its drawbacks. Its predictive performance is bounded by the resolution of the coarsest grid, which may limit its applicability in scenarios requiring high precision over large scales \cite{lai2024gm}. 


\paragraph{Point-SLAM}
Point-SLAM~\cite{sandström2023pointslam} introduces a dynamic neural point cloud approach that adapts density based on the detailed requirements driven by the data, significantly enhancing memory efficiency. By utilizing per-pixel image gradients, the model smartly modulates point density, concentrating on areas requiring more detail and simplifying regions with less complexity. As exploration progresses, Point-SLAM expands the point cloud, optimizing spatial usage by focusing point density and compressing points in less detailed regions, thereby maintaining computational efficiency in real-time scene reconstruction \cite{jiang2020dimensionality,jiang2024advanced,jiang2023enhancing}.

In Point-SLAM, the model's parameterization is crucial for fine-tuning its dynamic neural point cloud. Typical parameters include the resolution of the point cloud, which adjusts dynamically to match the density required by the scene complexity. The model also leverages parameters related to the gradient thresholds that control point density based on per-pixel image gradients. Additionally, parameters for space optimization ensure efficient memory use by managing how points are distributed and compressed in areas of lesser detail. These parameters are key to maintaining the balance between detail fidelity and computational efficiency in real-time applications.

\begin{figure}[h]
    \centering
    \includegraphics[width=0.5\textwidth]{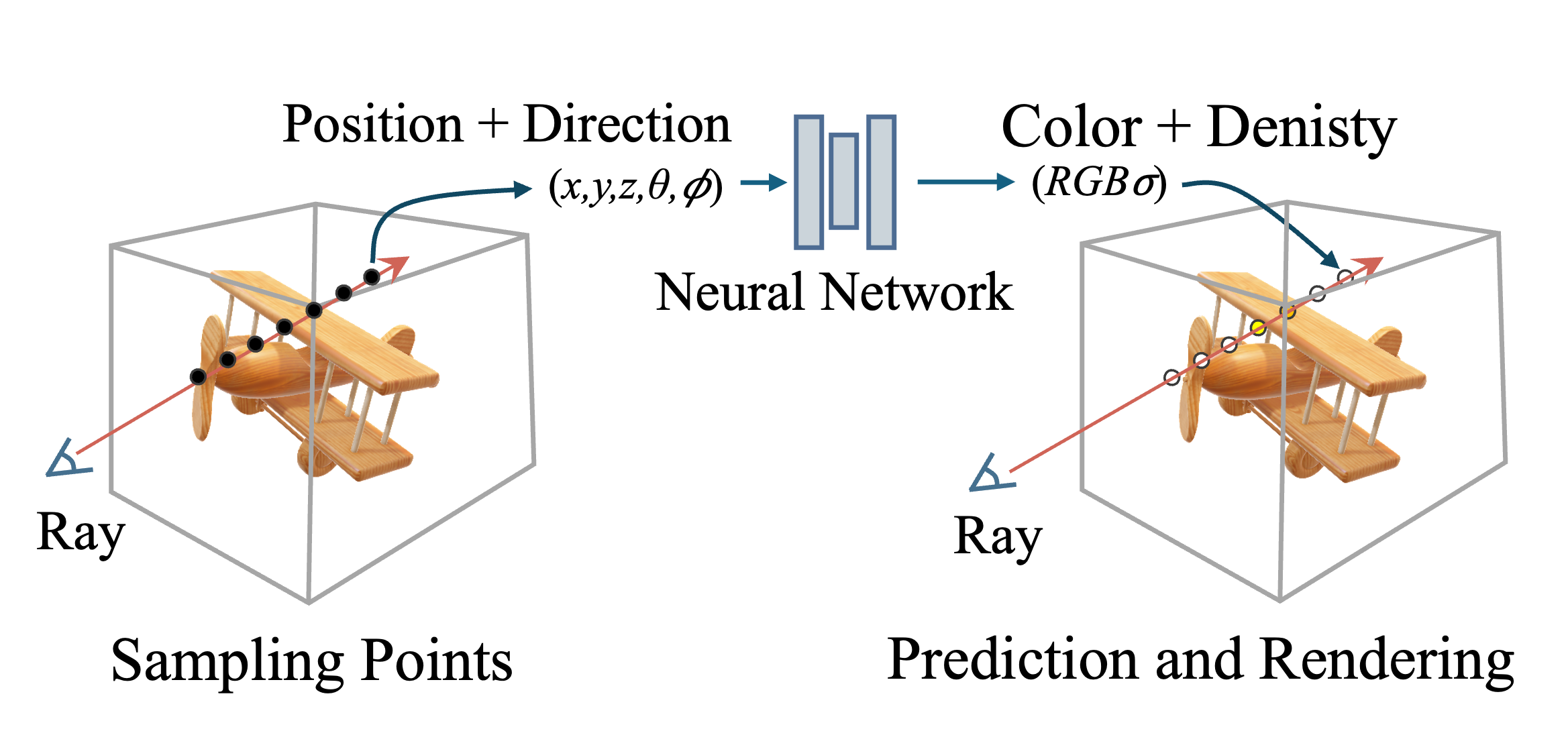} 
    \caption{High Level Architecture of NeRF Based 3D Reconstruction System}
    \label{fig:nerf}
\end{figure}

\subsubsection{Gaussian Based}

\paragraph{SplaTAM} SplaTAM \cite{keetha2024splatam} is a robust framework for dense RGB-D SLAM that utilizes advanced splatting techniques to achieve efficient tracking and mapping in 3D environments. SplaTAM leverages the representation of the environment using 3D Gaussian splats, which provide a continuous and smooth density function for effective integration of sensory data, enhancing the quality of mapping in complex scenes.  

The first step in SplaTAM involves data acquisition from RGB-D cameras, where both color and depth information are collected. The depth data is filtered and aligned with the RGB frames using methods from \cite{sturm2012benchmark}, ensuring consistency in representation. Subsequently, the algorithm constructs 3D Gaussians, known as "splats," from the incoming depth data. Each splat is represented by its mean, covariance, and color, enabling a rich representation of the scene's geometry and appearance \cite{durrant2006simultaneous}.  For tracking, SplaTAM employs an optimized keyframe selection technique, enabling significant computational savings while maintaining accurate localization. The framework utilizes a particle filter that tracks the agent's pose by propagating a set of hypotheses through the splat-based model. Each particle incorporates splat information, allowing for multi-hypothesis tracking, which is crucial in dynamic environments. The data association process employs a likelihood-based method, assessing the correspondence between observed features and existing splats to update particle weights \cite{moravec1980obstacle}. Mapping in SplaTAM is accomplished by integrating new observations into the existing splat representation in real-time using a Kalman filter approach, adjusting their parameters based on the incoming data. The algorithm intelligently manages occlusions and dynamic elements by classifying splats as either static or dynamic, utilizing robust outlier rejection strategies \cite{zhang2020robust} to ensure the integrity of the static map over time. 


Extensive evaluations demonstrate that SplaTAM outperforms traditional SLAM systems in terms of mapping fidelity and localization precision, making it suitable for real-time applications in challenging scenarios \cite{keetha2024splatam}. Metrics such as geometric consistency and computational efficiency are continuously monitored to ensure robust performance, providing an essential framework for navigating complex and dynamic environments. However, SplaTAM's computational complexity may hinder real-time performance, particularly in resource-constrained environments.

\paragraph{Gaussian Splatting SLAM}
Gaussian Splatting SLAM, introduced by Matsuki et al. \cite{matsuki2024gaussian}, aims to achieve real-time simultaneous localization and mapping using advanced Gaussian splatting techniques. The system integrates data from depth and RGB cameras, processing the information to model the environment while estimating the agent's trajectory accurately. 


The method represents each point in the 3D environment as a spherical Gaussian, defined by its mean and covariance, capturing local geometry and uncertainty. New observations are integrated into a global map by updating existing splats using a weighted average that accounts for measurement uncertainties, leading to smooth surface representations.  Keypoint detection algorithms like ORB or SIFT are employed to identify distinctive features, and a robust matching framework uses Gaussian properties to evaluate correspondences. RANSAC is applied to filter outliers, enhancing data association robustness, and loop closure detection improves mapping accuracy during long-duration operations \cite{kumlej2020loop,zhu2023pointclip}.


Gaussian Splatting SLAM is optimized for real-time operation, balancing efficient map updates with accurate localization to swiftly respond to environmental changes, which is crucial for robotic applications. Performance is evaluated through mapping accuracy, localization precision, and computational efficiency, assessing the quality of the generated map, the consistency of the agent's trajectory, and overall resource utilization \cite{matsuki2024gaussian}. Various optimization strategies, including graph-based optimization and bundle adjustment, ensure accurate estimates of both the agent's trajectory and the 3D map. However, the method may struggle with occlusions and dynamic environments, leading to inaccuracies, and it requires careful parameter tuning, which can limit adaptability across different platforms \cite{zhang2020robust}.

\begin{figure}[h]
    \centering
    \includegraphics[width=0.45\textwidth]{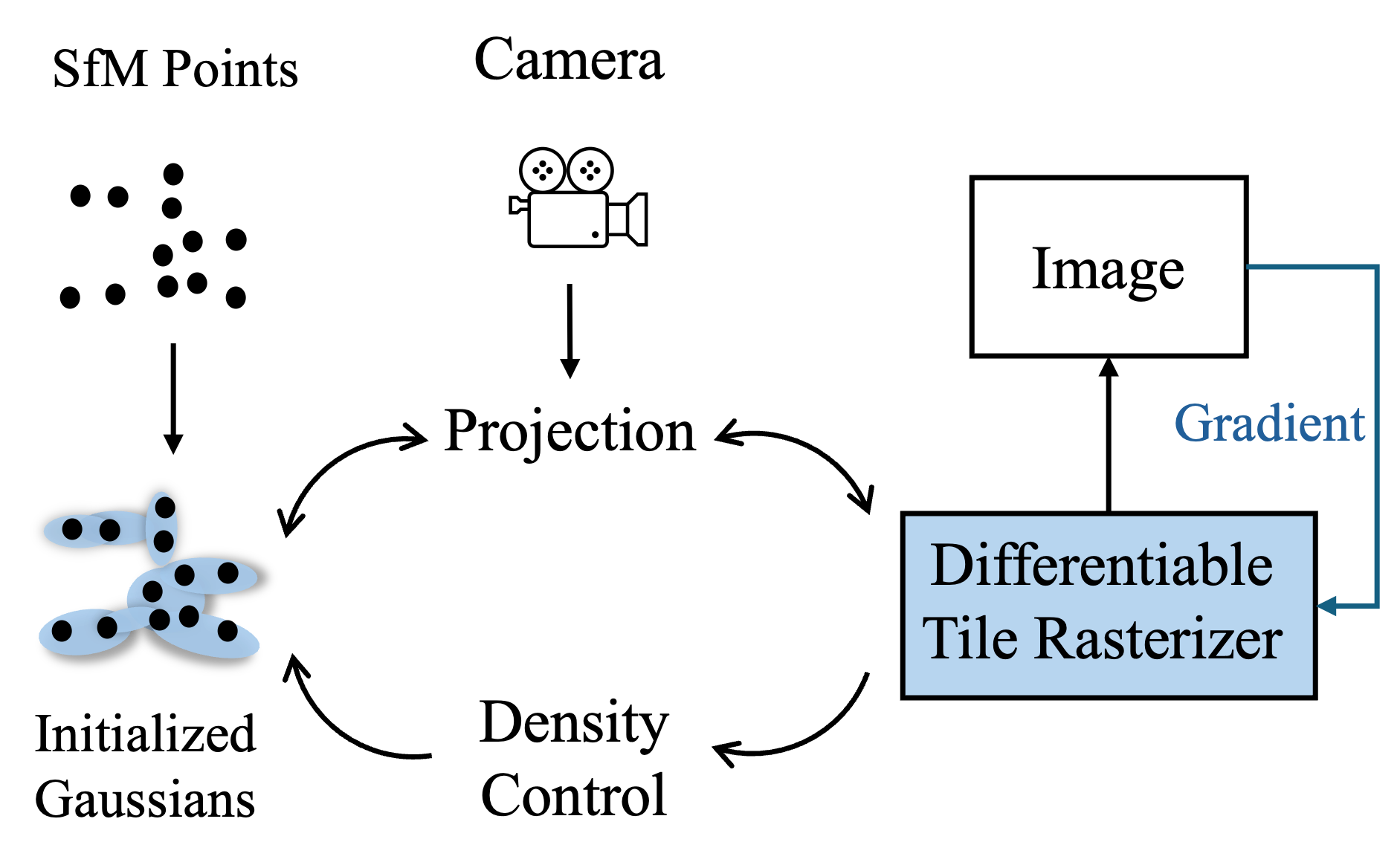} 
    \caption{Flowchart of 3D Gaussian Splatting}
    \label{fig:gs}
\end{figure}

\definecolor{LightGreen}{rgb}{0.85, 1, 0.85}
\definecolor{MediumGreen}{rgb}{0.6, 1, 0.6}
\definecolor{DarkGreen}{rgb}{0.2, 0.8, 0.2}
\definecolor{LightYellow}{rgb}{1, 1, 0.85}
\definecolor{MediumYellow}{rgb}{1, 1, 0.6}
\definecolor{DarkYellow}{rgb}{1, 1, 0.3}

\begin{table*}[ht]
    \centering
    \caption{\textbf{Rendering Performance on Replica.} Best rendering results are highlighted as \colorbox{DarkGreen!60}{first}.}
    \setlength{\tabcolsep}{3pt}
    \renewcommand{\arraystretch}{1.2}
    \begin{tabular}{l l c c c c c c c c c}
        \toprule
        \textbf{Method} & \textbf{Metric} & \textbf{Rm 0} & \textbf{Rm 1} & \textbf{Rm 2} & \textbf{Off 0} & \textbf{Off 1} & \textbf{Off 2} & \textbf{Off 3} & \textbf{Off 4} & \textbf{Avg.} \\
        \midrule
        \multirow{4}{*}{\textbf{NICE-SLAM \cite{zhu2022niceslam}}} & PSNR [dB] $\uparrow$ & 22.12 & 22.47 & 24.52 & 29.07 & 30.34 & 19.66 & 22.23 & 24.94 & 24.42 \\
        & SSIM $\uparrow$ & 0.69 & 0.76 & 0.81 & 0.87 & 0.89 & 0.80 & 0.80 & 0.86 & 0.81 \\
        & LPIPS $\downarrow$ &  0.33 &  0.27 &  0.21 &  0.23 &  0.18 &  0.24 &  0.21 & 0.20 &  0.23 \\
        
        \midrule
        \multirow{4}{*}{\textbf{Point-SLAM \cite{sandström2023pointslam}}} & PSNR [dB] $\uparrow$ & 32.40 & 34.08 & 35.50 & 38.26 & 39.16 & 33.99 & \ 33.48 & 33.49 & 35.17 \\
        & SSIM $\uparrow$ & 0.97 & \cellcolor{DarkGreen!60} 0.98 & \cellcolor{DarkGreen!60} 0.98 & \cellcolor{DarkGreen!60} 0.98 & \cellcolor{DarkGreen!60} 0.99 & 0.96 & \cellcolor{DarkGreen!60} 0.96 & \cellcolor{DarkGreen!60} 0.98 & \cellcolor{DarkGreen!60} 0.98 \\
        & LPIPS $\downarrow$ & 0.12 & 0.11 & 0.12 & 0.11 & 0.10 & 0.12 & 0.16 & 0.13 & 0.14 \\

        \midrule
        \multirow{4}{*}{\textbf{SplaTAM \cite{keetha2024splatam}}} & PSNR [dB] $\uparrow$ & 32.86 & 33.89 & 35.25 & 38.26 & 39.17 & 31.97 & 29.70 & 31.81 & 34.11 \\
        & SSIM $\uparrow$ & \cellcolor{DarkGreen!60} 0.98 & 0.97 & \cellcolor{DarkGreen!60} 0.98 & \cellcolor{DarkGreen!60} 0.98 & 0.98 & \cellcolor{DarkGreen!60} 0.97 & 0.95 & 0.95 & 0.97 \\
        & LPIPS $\downarrow$ & 0.07 & 0.10 & 0.08 & 0.09 & 0.09 & 0.10 & 0.12 & 0.15 & 0.10 \\

        \midrule
        \multirow{4}{*}{\textbf{Gaussian Splatting SLAM \cite{matsuki2024gaussian}}} & PSNR [dB] $\uparrow$ & \cellcolor{DarkGreen!60} 34.83 & \cellcolor{DarkGreen!60} 36.43 &  \cellcolor{DarkGreen!60} 37.49 & \cellcolor{DarkGreen!60} 39.95 & \cellcolor{DarkGreen!60} 42.09 & \cellcolor{DarkGreen!60} 36.24 & \cellcolor{DarkGreen!60} 36.70 & \cellcolor{DarkGreen!60} 36.07 & \cellcolor{DarkGreen!60} 37.50 \\
        & SSIM $\uparrow$ & 0.95 & 0.96 & 0.97 & 0.97 & 0.98 & \cellcolor{DarkGreen!60} 0.97 & \cellcolor{DarkGreen!60} 0.96 & 0.96 & 0.96 \\
        & LPIPS $\downarrow$ & \cellcolor{DarkGreen!60} 0.07 & \cellcolor{DarkGreen!60} 0.07 & \cellcolor{DarkGreen!60} 0.08 & \cellcolor{DarkGreen!60} 0.07 & \cellcolor{DarkGreen!60} 0.06 & \cellcolor{DarkGreen!60} 0.08 & \cellcolor{DarkGreen!60} 0.07 & \cellcolor{DarkGreen!60} 0.10 & \cellcolor{DarkGreen!60} 0.07 \\
        
        \bottomrule
    \end{tabular}
    \label{tab:rendering}
\end{table*}

\begin{table*}[htbp]
    \centering
    \caption{\textbf{Tracking Performance on Replica (ATE RMSE ↓ [cm])} Best results are highlighted as \colorbox{DarkGreen!60}{first}, \colorbox{MediumGreen!60}{second} and \colorbox{DarkYellow!60}{third}}
    \setlength{\tabcolsep}{12pt}
    \renewcommand{\arraystretch}{1.0}
    \resizebox{0.8\textwidth}{!}{
    \begin{tabular}{l l c c c c c c c c c}
        \toprule
        \textbf{Methods}  & \textbf{R0} & \textbf{R1} & \textbf{R2} & \textbf{Of0} & \textbf{Of1} & \textbf{Of2} & \textbf{Of3} & \textbf{Of4} & \textbf{Avg.}\\
        \midrule
        NICE-SLAM \cite{zhu2022niceslam}  & 0.97 & 1.31 & 1.07 & 0.88 & 1.00 & 1.06 & 1.10 & 1.13 & 1.06\\
        Point-SLAM \cite{sandström2023pointslam}  & \cellcolor{DarkYellow!60} 0.61 & \cellcolor{DarkYellow!60} 0.41 & \cellcolor{MediumGreen!60} 0.37 & \cellcolor{MediumGreen!60} 0.38 & \cellcolor{DarkYellow!60} 0.48 & \cellcolor{DarkYellow!60} 0.54 & \cellcolor{MediumGreen!60} 0.69 & \cellcolor{DarkYellow!60} 0.72 & \cellcolor{DarkYellow!60} 0.52\\
        
        SplaTAM \cite{keetha2024splatam}  & \cellcolor{DarkGreen!60} 0.31 & \cellcolor{MediumGreen!60} 0.40 & \cellcolor{DarkGreen!60} 0.29 & \cellcolor{DarkGreen!60} 0.47 & \cellcolor{MediumGreen!60} 0.27 & \cellcolor{MediumGreen!60} 0.29 & \cellcolor{DarkGreen!60} 0.32 & \cellcolor{DarkGreen!60} 0.55 & \cellcolor{DarkGreen!60} 0.36\\
        
        Gaussian Splatting SLAM \cite{matsuki2024gaussian}  & \cellcolor{MediumGreen!60} 0.32 & \cellcolor{DarkGreen!60} 0.31 & \cellcolor{DarkYellow!60}  0.44 & \cellcolor{DarkYellow!60} 0.52 & \cellcolor{DarkGreen!60} 0.23 & \cellcolor{DarkGreen!60} 0.17 & \cellcolor{DarkYellow!60} 2.25 &\cellcolor{MediumGreen!60} 0.58 & \cellcolor{MediumGreen!60} 0.44\\
        
        \bottomrule
    \end{tabular}}
    \label{tab:tracking}
\end{table*}

\begin{table*}[htbp]
    \centering
    \caption{\textbf{Tracking Performance on ScanNet (ATE RMSE ↓ [cm])}}
    \setlength{\tabcolsep}{12pt}
    \renewcommand{\arraystretch}{1.0}
    \resizebox{0.8\textwidth}{!}{
    \begin{tabular}{l l c c c c c c c c c}
        \toprule
        \textbf{Methods}  & \textbf{0000} & \textbf{0059} & \textbf{0106} & \textbf{0169} & \textbf{0181} & \textbf{0207} & \textbf{Avg.} \\
        \midrule
        NICE-SLAM \cite{zhu2022niceslam}  & \cellcolor{DarkGreen!60} 8.64 & 12.25 & \cellcolor{DarkGreen!60} 8.09 & \cellcolor{DarkGreen!60} 10.28 & \cellcolor{DarkYellow!60} 12.93 & \cellcolor{DarkGreen!60} 5.59 & \cellcolor{DarkGreen!60} 9.63 \\ 
        
        Point-SLAM \cite{sandström2023pointslam}  & \cellcolor{MediumGreen!60} 10.24 & \cellcolor{DarkGreen!60} 7.81 & \cellcolor{MediumGreen!60} 8.65 & 22.16 & 14.77 & 9.54 & 12.19 \\
        
        SplaTAM \cite{keetha2024splatam}  & 12.83 & \cellcolor{DarkYellow!60} 10.10 & 17.72 & \cellcolor{DarkYellow!60} 12.08 & \cellcolor{MediumGreen!60} 11.10 & \cellcolor{DarkYellow!60} 7.46 & \cellcolor{DarkYellow!60} 11.88\\

        Gaussian Splatting SLAM \cite{matsuki2024gaussian} & \cellcolor{DarkYellow!60} 11.28 & \cellcolor{MediumGreen!60} 9.24 & \cellcolor{DarkYellow!60} 16.49 & \cellcolor{MediumGreen!60} 11.09 & \cellcolor{DarkGreen!60} 10.88 & \cellcolor{MediumGreen!60} 6.58 & \cellcolor{MediumGreen!60} 10.84 \\

        \bottomrule
    \end{tabular}}
    \label{tab:tracking}
\end{table*}

\begin{figure*}[htbp]
    \centering
    \caption{Reconstruction performance on Replica \cite{straub2019replica} using \textbf{NeRF-Based} methods. The columns display reconstruction results from NICE-SLAM and Point-SLAM, compared to the Ground Truth.}

    \begin{minipage}{0.33\textwidth}
        \centering
        \includegraphics[width=0.8\textwidth]{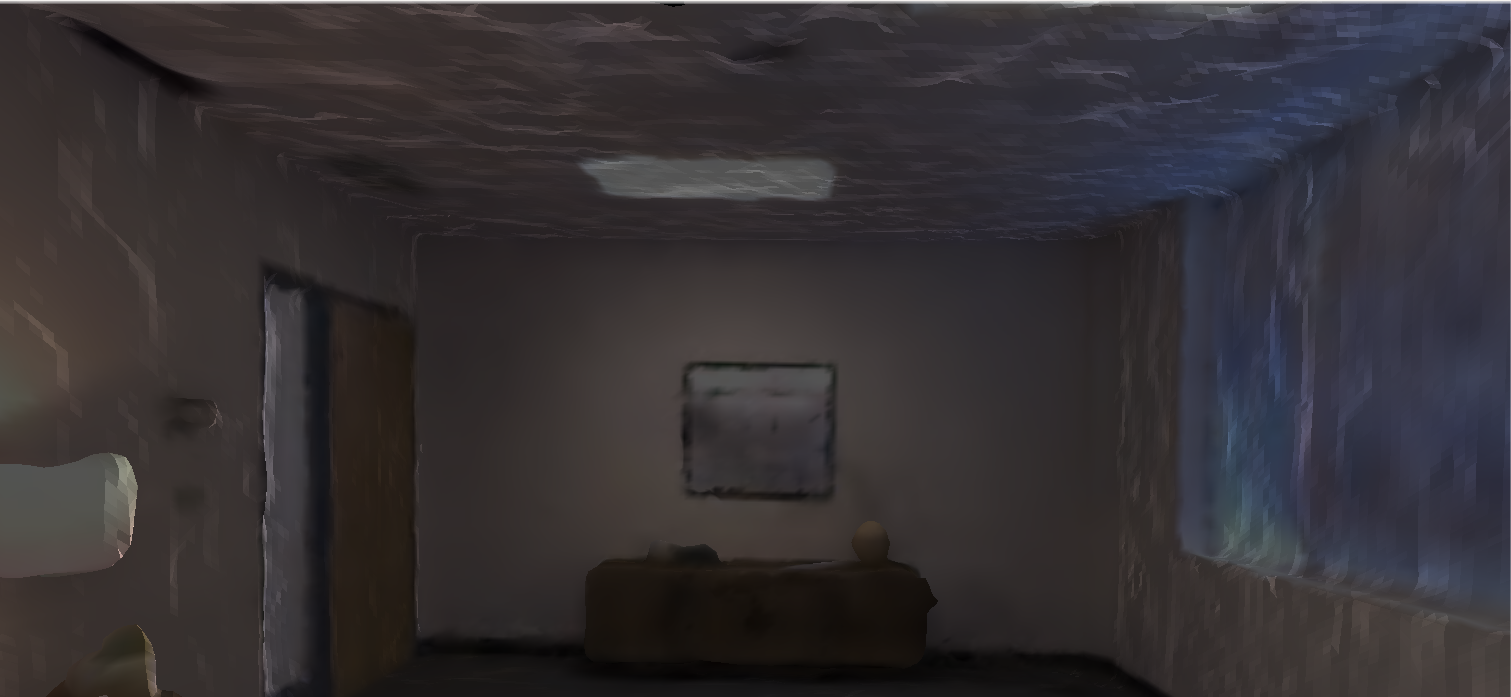}\\[1mm]
        \includegraphics[width=0.8\textwidth]{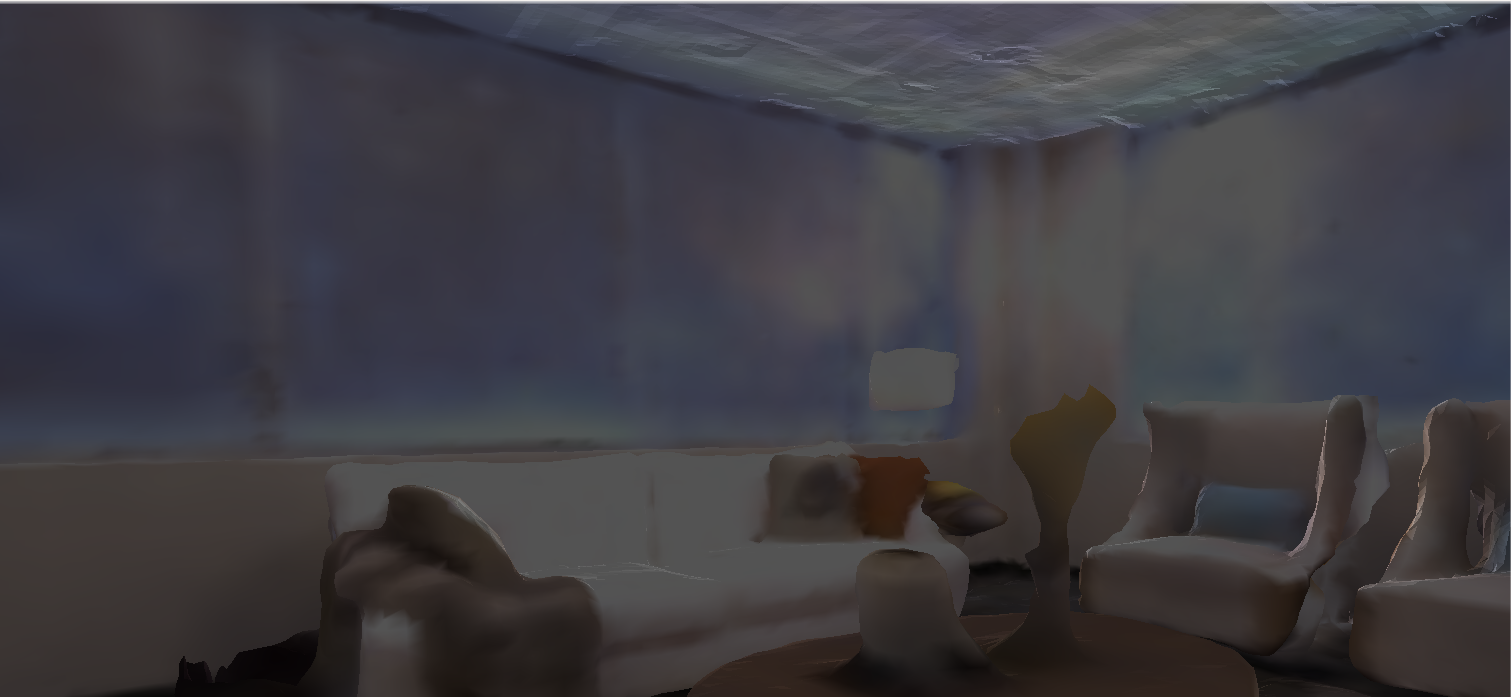}\\[1mm]
        \includegraphics[width=0.8\textwidth]{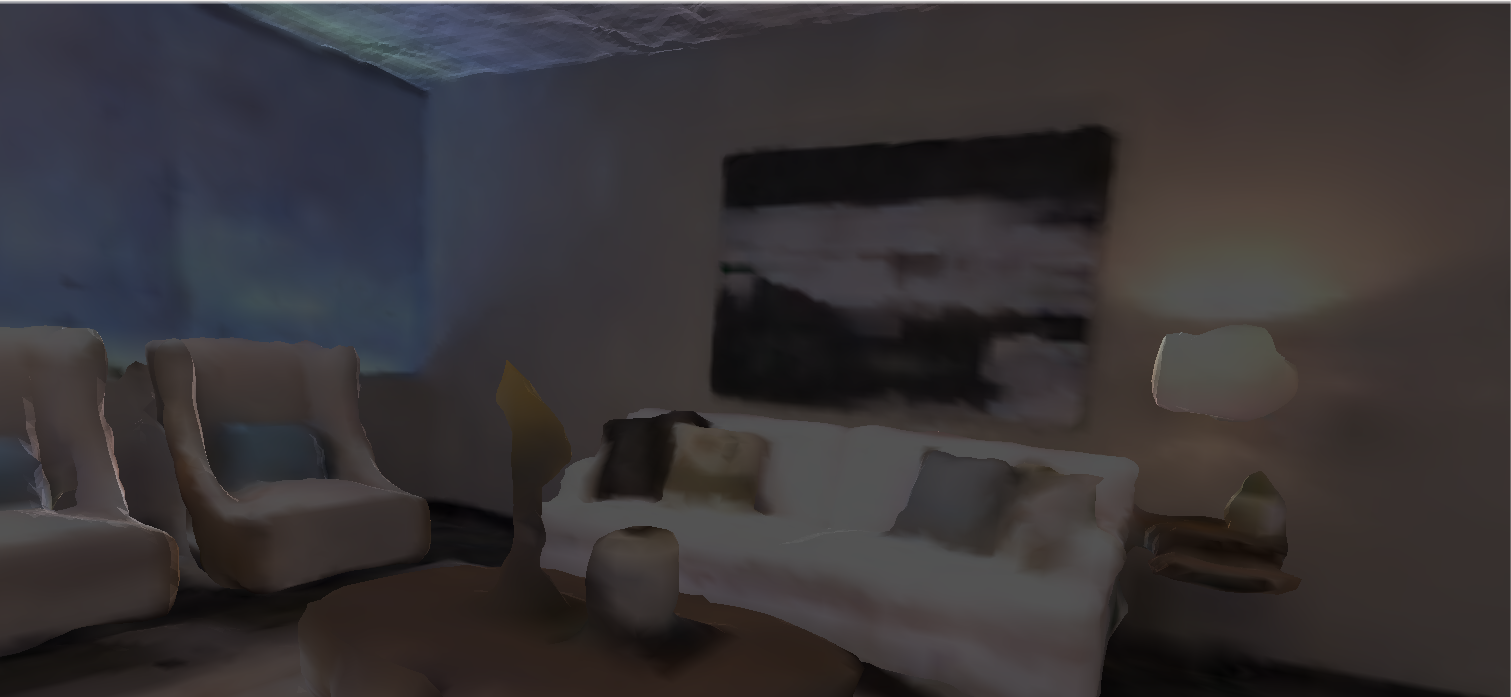}
    \end{minipage}\hfill
    \begin{minipage}{0.33\textwidth}
        \centering
        \includegraphics[width=0.8\textwidth]{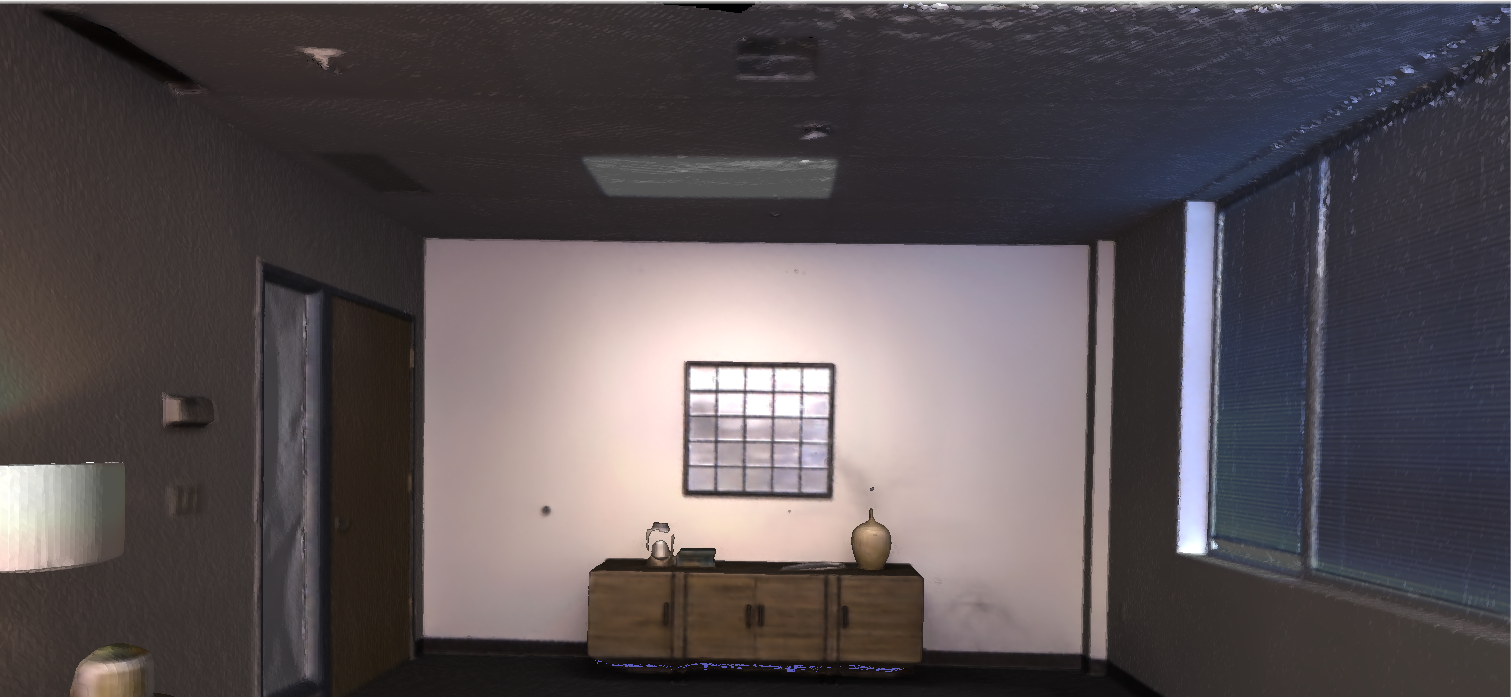}\\[1mm]
        \includegraphics[width=0.8\textwidth]{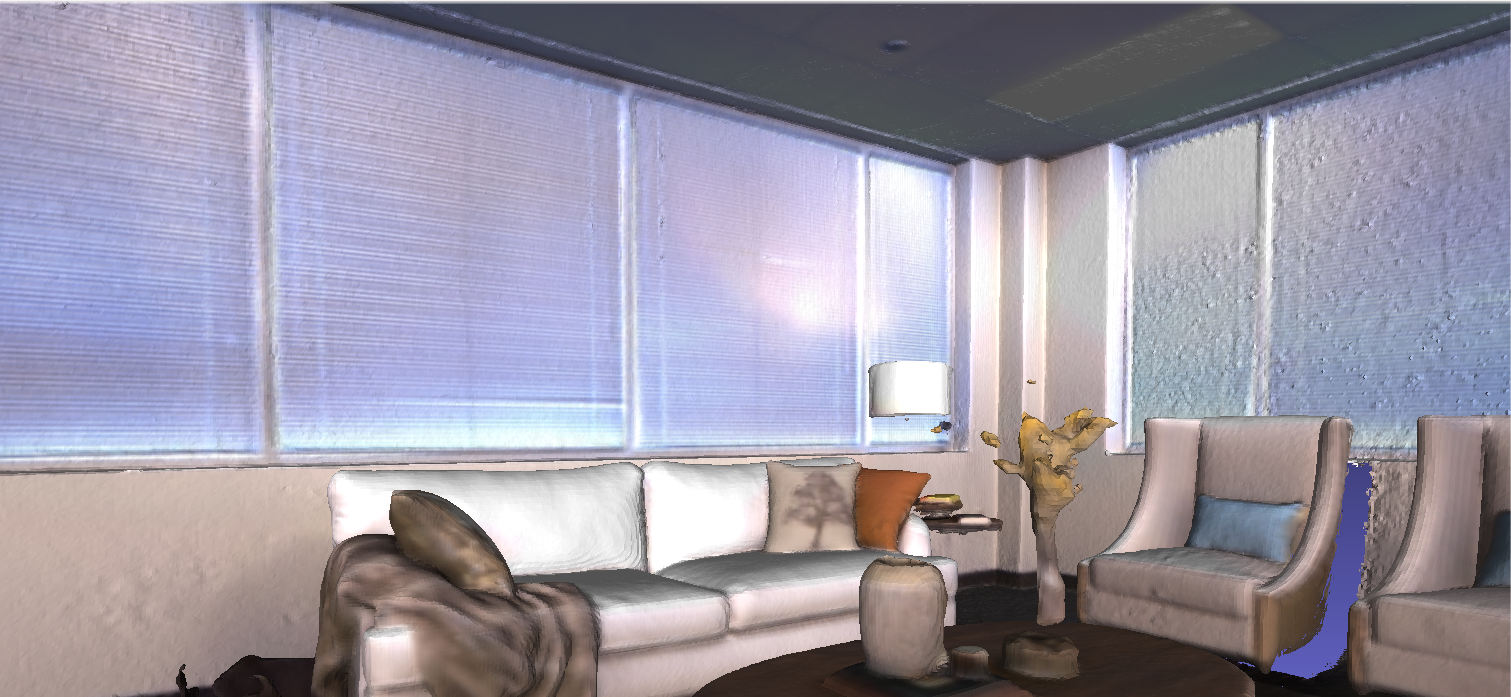}\\[1mm]
        \includegraphics[width=0.8\textwidth]{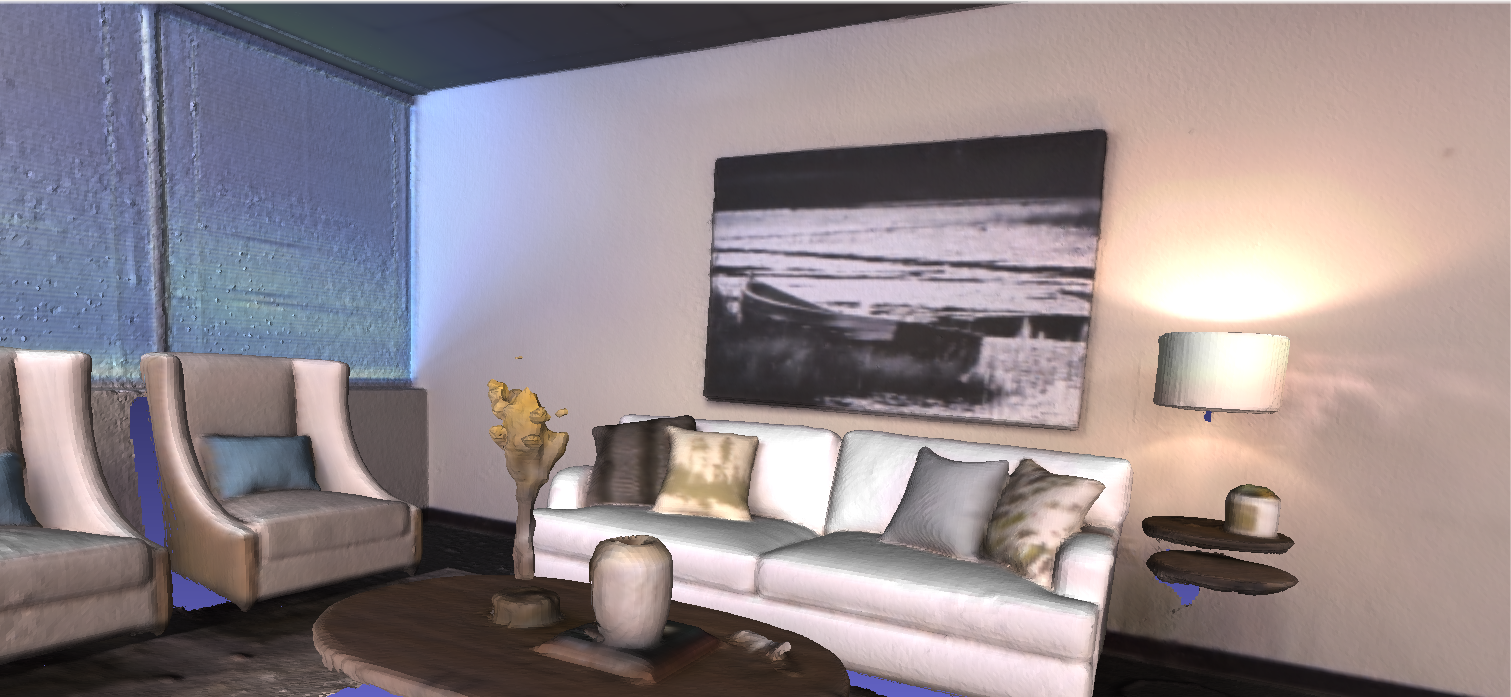}
    \end{minipage}\hfill
    \begin{minipage}{0.33\textwidth}
        \centering
        \includegraphics[width=0.8\textwidth]{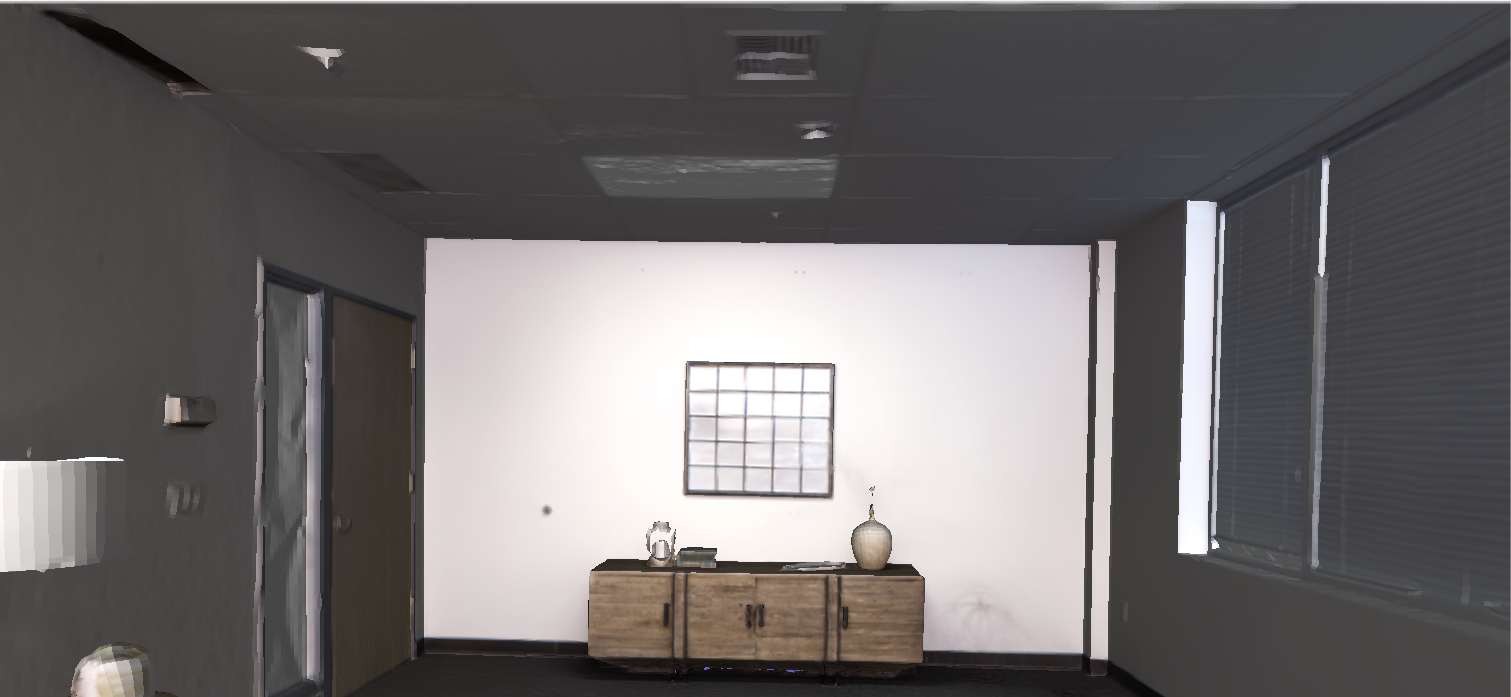}\\[1mm]
        \includegraphics[width=0.8\textwidth]{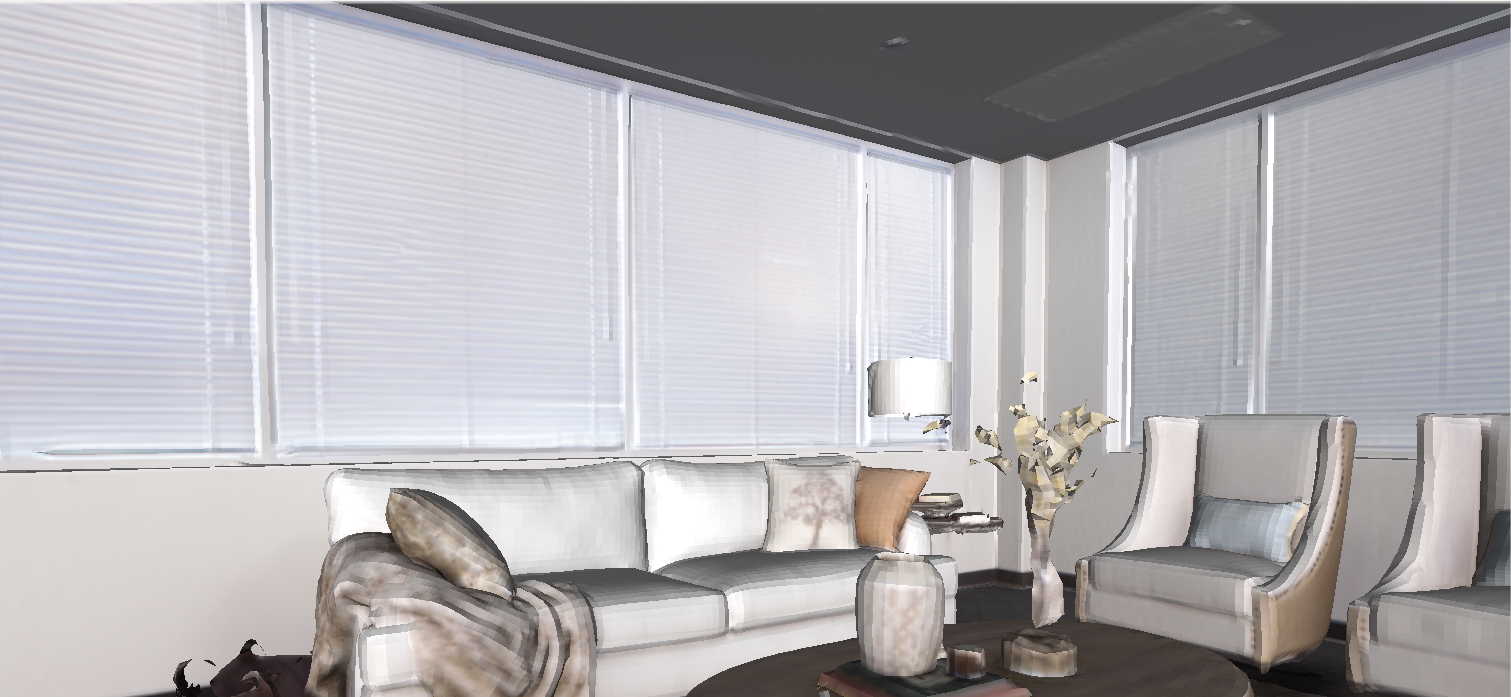}\\[1mm]
        \includegraphics[width=0.8\textwidth]{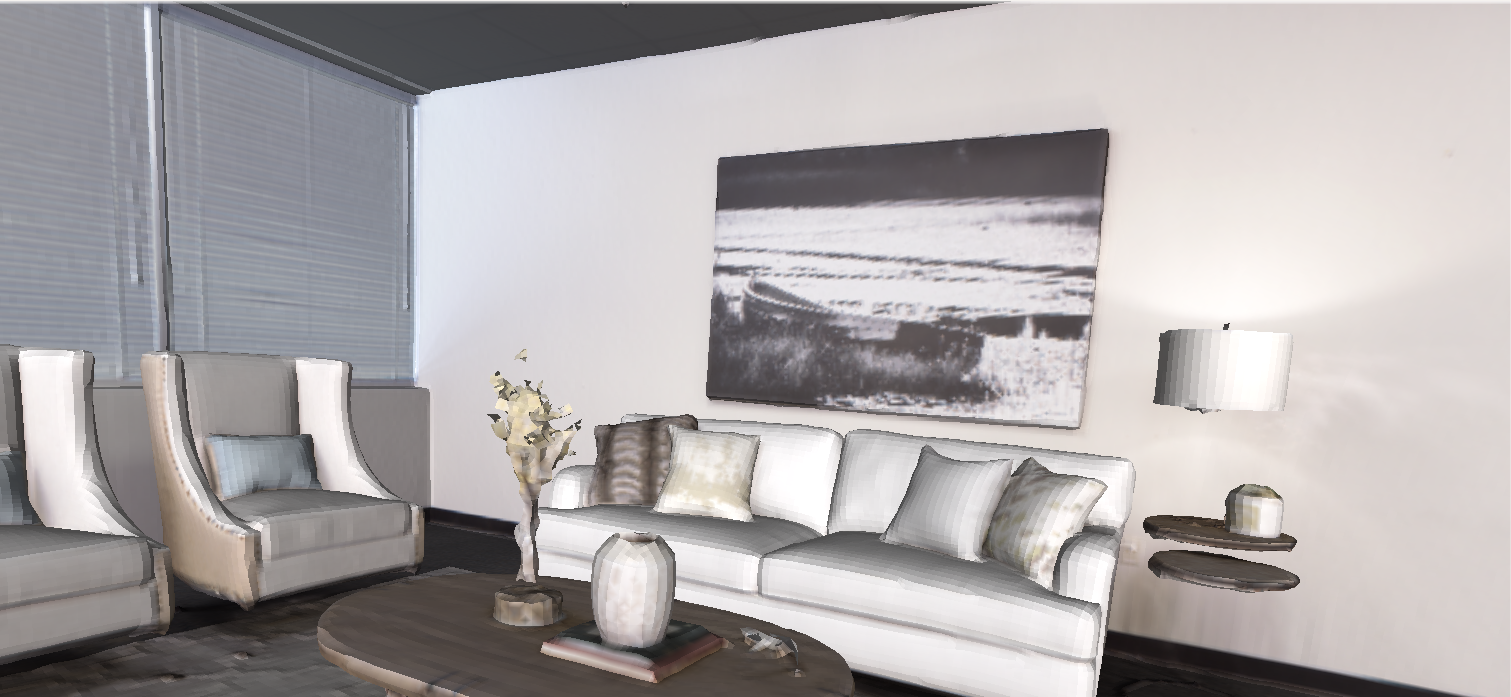}
    \end{minipage}
    
    \begin{minipage}{0.33\textwidth}
        \centering
        \subcaption{NICE-SLAM \cite{zhu2022niceslam}}
    \end{minipage}\hfill
    \begin{minipage}{0.33\textwidth}
        \centering
        \subcaption{Point-SLAM \cite{sandström2023pointslam}}
    \end{minipage}\hfill
    \begin{minipage}{0.33\textwidth}
        \centering
        \subcaption{Ground Truth}
    \end{minipage}
\end{figure*}

\begin{figure*}[htbp]
    \centering
    \caption{Reconstruction performance on Replica \cite{straub2019replica} using \textbf{Gaussian Splatting Based} methods. The top row shows results from SplaTAM, and the bottom row shows results from Gaussian Splatting SLAM. The first two columns display interior views, while the last column shows the exterior view. The GS-based methods have difficulty creating smooth surfaces, leading to noticeable Gaussian bubbles compared to NeRF-based methods.}

    \begin{minipage}{0.33\textwidth}
        \centering
        \subcaption{SplaTAM \cite{keetha2024splatam}}
    \end{minipage}\\[-2.6mm]
    
    \begin{minipage}{0.33\textwidth}
        \centering
        \rotatebox{180}{\includegraphics[width=4.8cm, height=2.5cm]{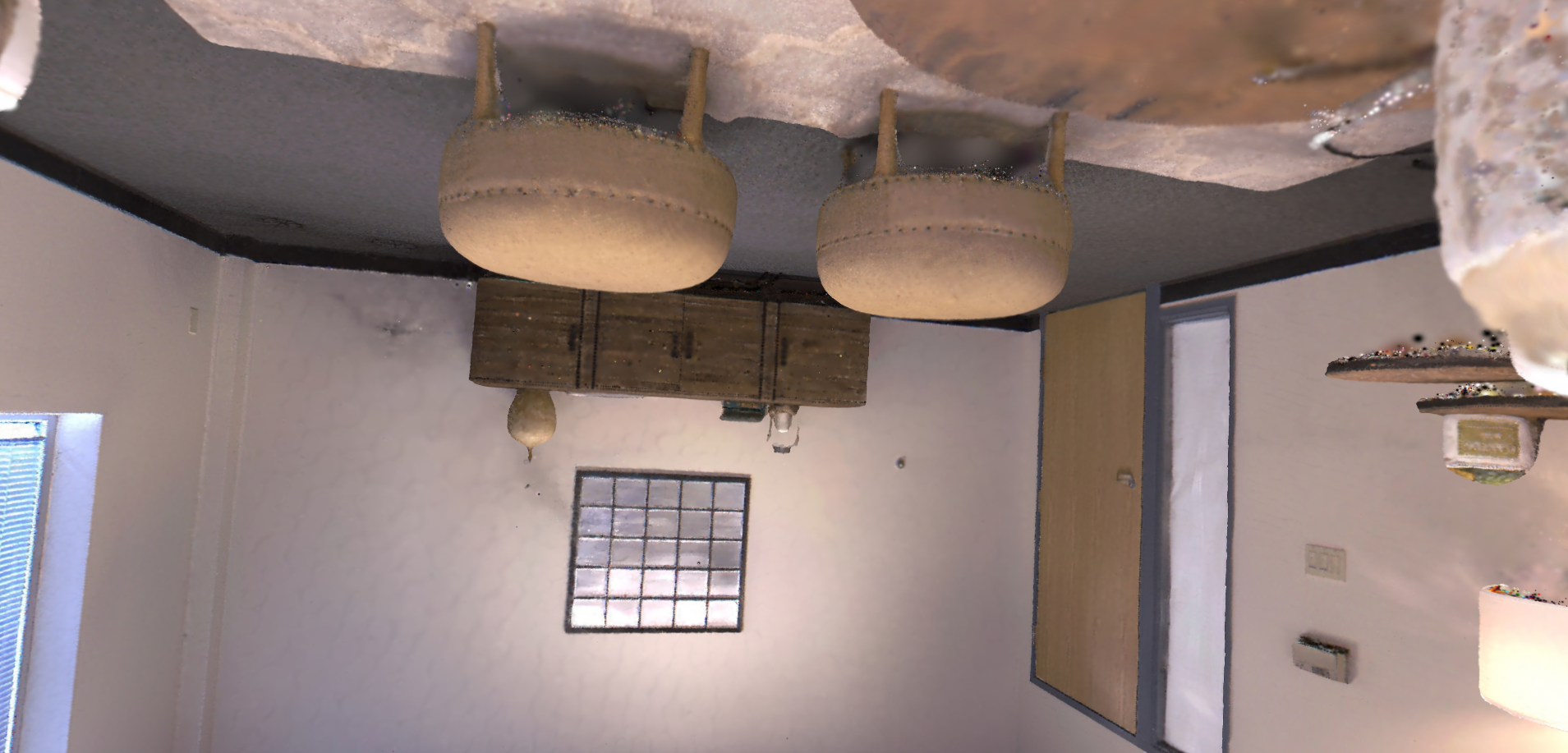}}\\[1mm]
        \rotatebox{0}{\includegraphics[width=4.8cm, height=2.5cm]{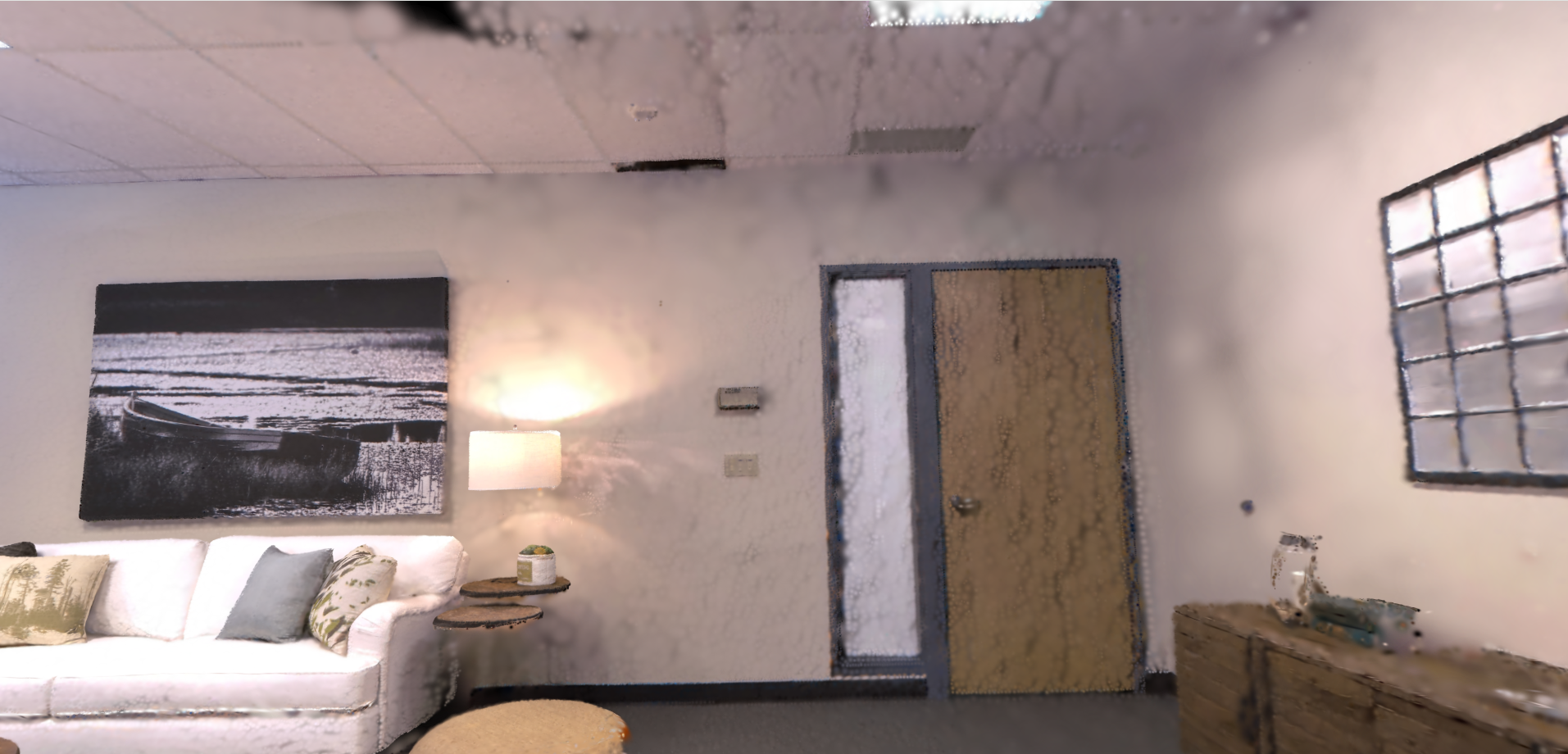}}\\[1mm]
    \end{minipage}\hfill
    \begin{minipage}{0.33\textwidth}
        \centering
        \rotatebox{180}{\includegraphics[width=4.8cm, height=2.5cm]{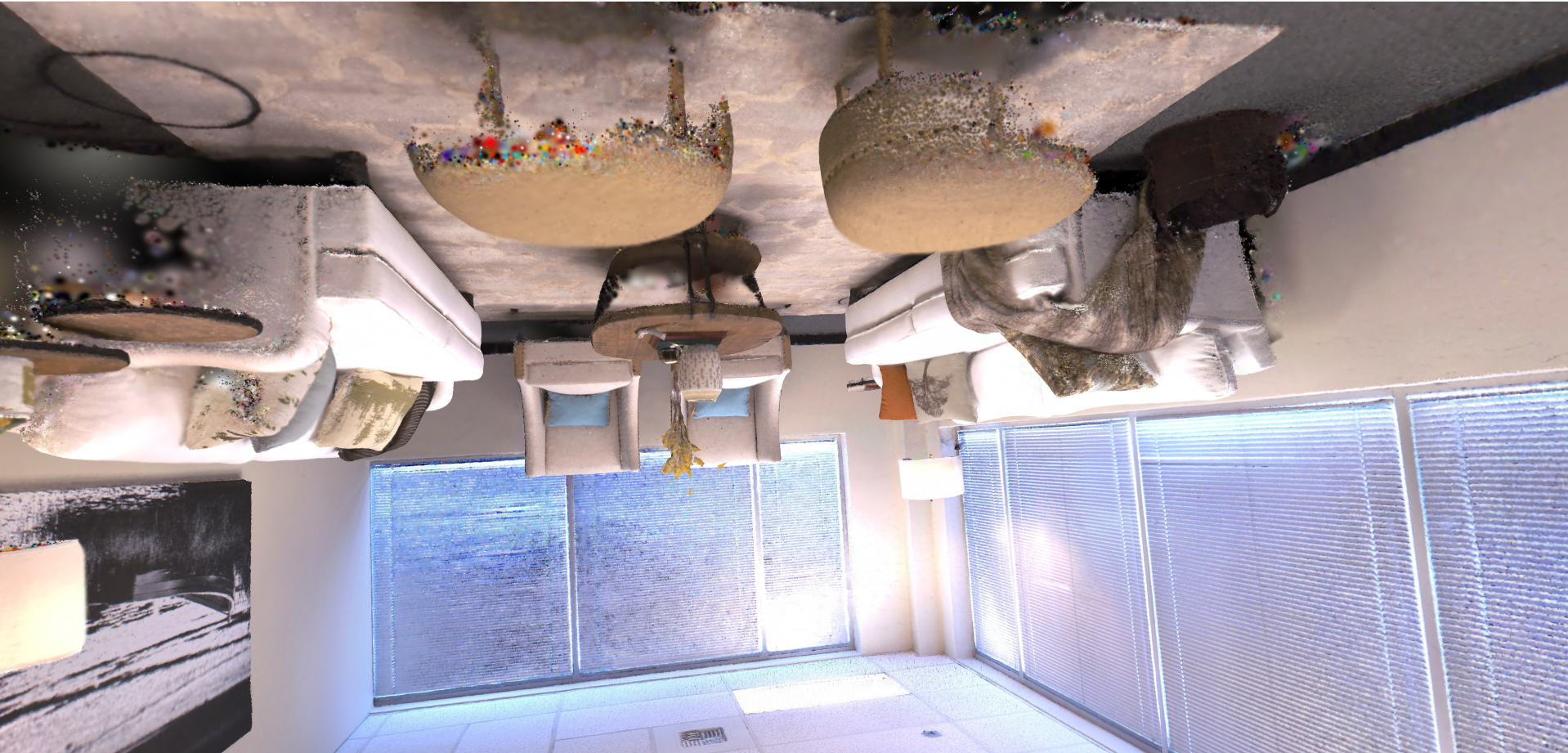}}\\[1mm]
        \rotatebox{0}{\includegraphics[width=4.8cm, height=2.5cm]{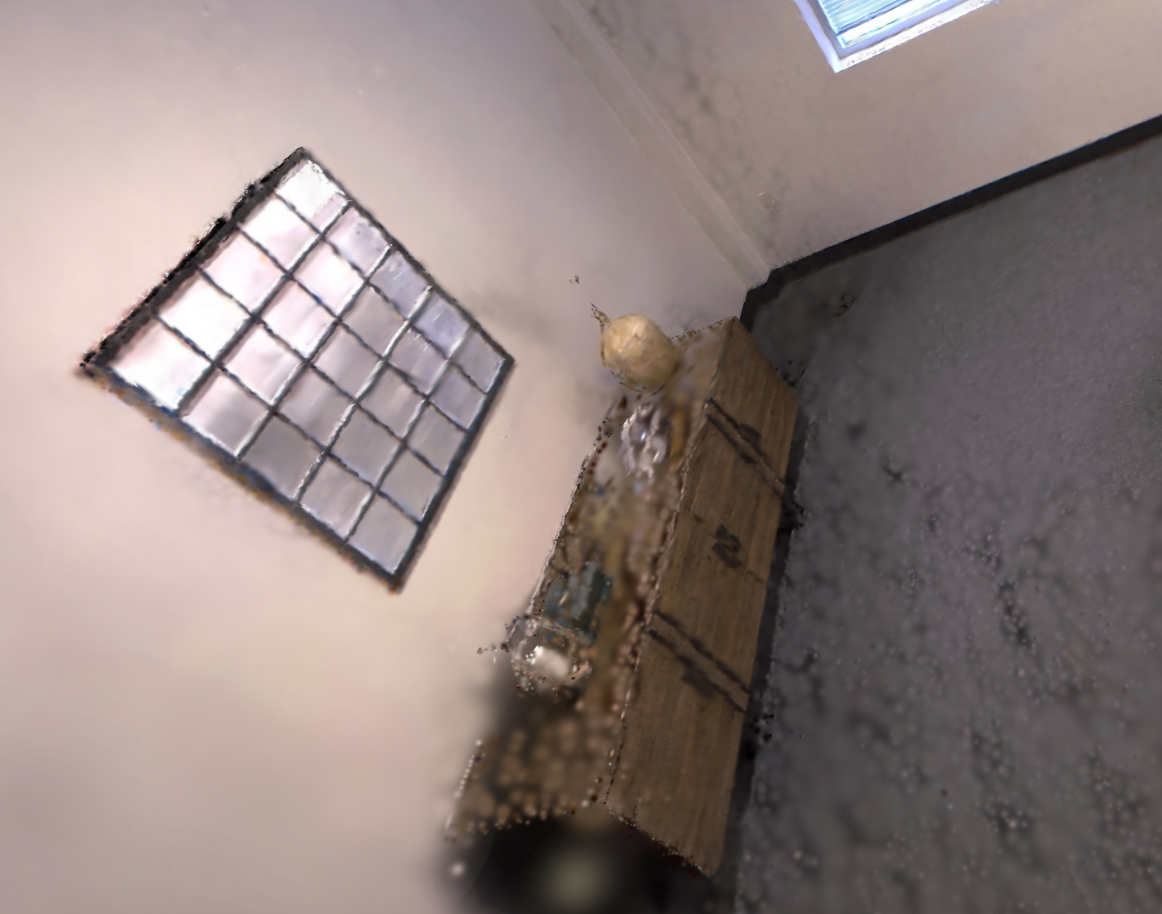}}\\[1mm]
    \end{minipage}\hfill
        \begin{minipage}{0.33\textwidth}
        \centering
        \rotatebox{0}{\includegraphics[width=4.8cm, height=2.5cm]{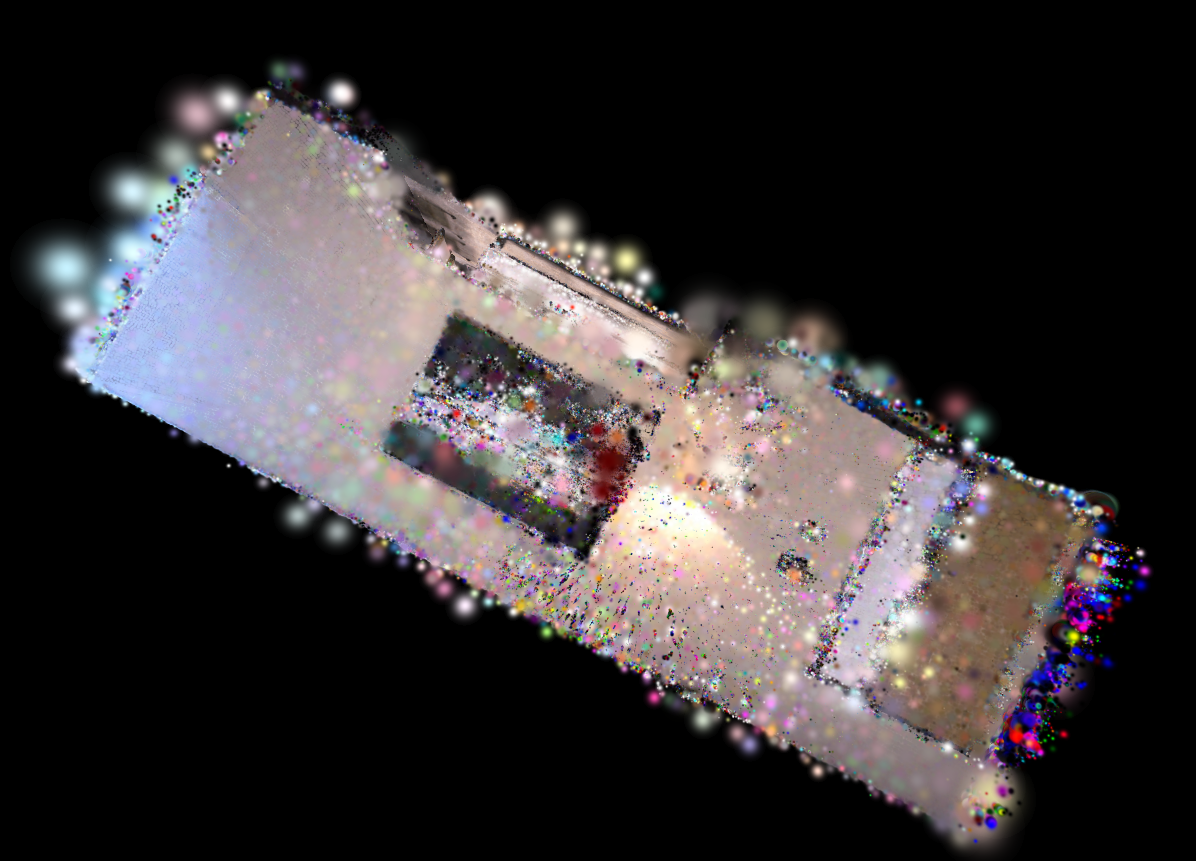}}\\[1mm]
        \rotatebox{0}{\includegraphics[width=4.8cm, height=2.5cm]{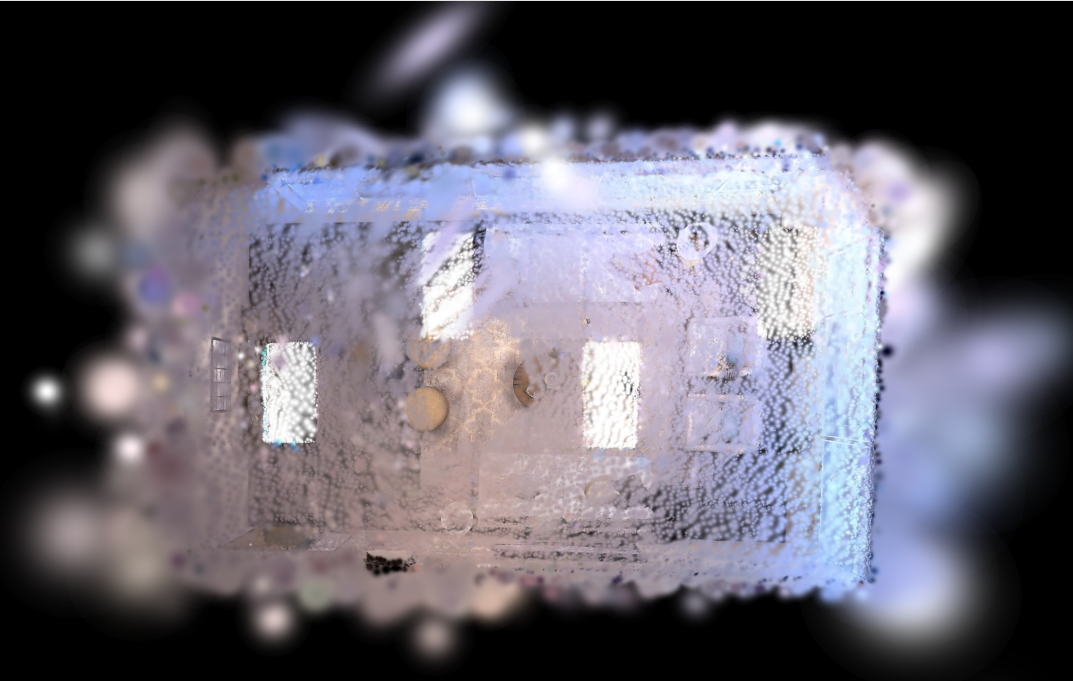}}\\[1mm]
    \end{minipage}

    \begin{minipage}{0.33\textwidth}
        \centering
        \subcaption{Gaussian Splatting SLAM \cite{matsuki2024gaussian}}
    \end{minipage}
\end{figure*}

\section{Evaluation}

\subsection{Evaluation Metrics}
In the context of 3D reconstruction, the performance is rigorously evaluated across three primary domains: tracking, mapping, and view synthesis. Tracking assesses the system's ability to accurately follow the trajectory of the camera through space, crucial for maintaining alignment with the 3D model. Mapping focuses on the accuracy and completeness of the 3D structure generated from the captured data, emphasizing the need for precise and comprehensive environmental details. View synthesis, on the other hand, examines the system's capability to create new viewpoints from the reconstructed 3D model, testing the model's utility for applications like virtual reality, where users may explore the environment from angles not originally captured by the data. Each domain leverages specific metrics tailored to assess distinct aspects of the reconstruction process, ensuring a holistic evaluation of the system’s performance. Furthermore, leveraging Zhang’s method~\cite{zhang2024deepgi}, our proposal incorporates adaptive learning techniques to further enhance the data analysis process, aiming to optimize outcomes in future planning.

\subsubsection{Tracking}
\begin{itemize}
    \item Absolute Trajectory Error (ATE): Measures the discrepancy between the estimated camera trajectory and the ground truth trajectory. This metric is essential for assessing the precision with which the SLAM system can track the movement through the environment.
\end{itemize}

\subsubsection{Mapping}
\begin{itemize}
    \item Precision: Indicates the accuracy of the points that are positively identified as part of the structure. It assesses how many reconstructed points are true positives.
    \item Recall: Measures the algorithm’s ability to identify all relevant instances of the structure in the scene. It evaluates the completeness of the reconstruction.
    \item Depth L1 Error: Calculates the absolute difference between the predicted depth values and the actual depth values in the ground truth. This metric is crucial for understanding the depth accuracy of the model.
\end{itemize}

\subsubsection{View Synthesis}
\begin{itemize}
    \item Peak Signal to Noise Ratio (PSNR): Evaluates the quality of reconstructed views by comparing the synthesized images against the original views, focusing on the fidelity and clarity of the visual outputs.
\end{itemize}

\subsection{Comparative Analysis}
This section evaluates and compares the performance of four state-of-the-art 3D scene reconstruction algorithms: NICE-SLAM, Point-SLAM, SplaTAM, and Gaussian Splatting SLAM, utilizing two main performance metrics: rendering quality and tracking accuracy. The data from these evaluations are derived from experiments conducted on the Replica dataset.

\subsubsection{Rendering Performance}

Rendering performance is quantified through three metrics: PSNR (Peak Signal-to-Noise Ratio), SSIM (Structural Similarity Index), and LPIPS (Learned Perceptual Image Patch Similarity). Higher values in PSNR and SSIM indicate better image reconstruction quality, while a lower LPIPS suggests a closer perceptual resemblance to the ground truth.

\begin{itemize}
    \item \textbf{NICE-SLAM} shows moderate performance with average PSNR, SSIM, and LPIPS values of 24.42 dB, 0.81, and 0.23 respectively. It performs particularly well in office scenarios but struggles in highly textured environments, indicating a potential limitation in handling complex textures or lighting conditions.
    \item \textbf{Point-SLAM} exhibits superior rendering quality with the highest average PSNR and SSIM scores of 35.17 dB and 0.98, respectively, and a low LPIPS of 0.14. Its performance highlights its robustness in both detailed and varied environmental conditions.
    \item \textbf{SplaTAM} also demonstrates strong rendering capabilities, with a comparable PSNR to Point-SLAM at 34.11 dB and a slightly lower SSIM of 0.97. Its average LPIPS of 0.10 signifies a high perceptual quality, making it effective in diverse settings.
    \item \textbf{Gaussian Splatting SLAM} leads in PSNR at 37.50 dB and shows a good balance in SSIM (0.96) and LPIPS (0.07), reflecting its efficiency in creating high-fidelity reconstructions across different scenes.
\end{itemize}

\subsubsection{Tracking Performance}

Tracking accuracy is evaluated using the ATE RMSE metric, where lower values indicate more precise trajectory tracking.

\begin{itemize}
    \item \textbf{NICE-SLAM} records an average ATE RMSE of 1.06 cm, showing reliable tracking but with some inconsistencies across different environments.
    \item \textbf{Point-SLAM}, with an average ATE RMSE of 0.52 cm, demonstrates exceptional tracking precision, outperforming other models particularly in complex indoor environments.
    \item \textbf{SplaTAM} achieves the best tracking performance with an impressively low average ATE RMSE of 0.36 cm, indicating highly accurate trajectory estimation.
    \item \textbf{Gaussian Splatting SLAM} shows mixed results in tracking, with an overall average ATE RMSE of 0.44 cm. It performs exceptionally well in less dynamic environments but exhibits significant errors in more challenging settings, as indicated by the outlier value of 2.25 cm in one of the office scenes.
\end{itemize}

\subsubsection{Overall Assessment}

When considering both rendering and tracking performances, Point-SLAM and SplaTAM emerge as leaders, offering robust solutions for 3D scene reconstruction with high fidelity and precision. NICE-SLAM, while effective, shows limitations under complex conditions. Gaussian Splatting SLAM provides excellent rendering results but faces challenges in maintaining consistent tracking accuracy across all scenarios. 


\section{Conclusions}
This study provides a comprehensive evaluation of four leading 3D scene reconstruction algorithms: NICE-SLAM, Point-SLAM, SplaTAM, and Gaussian Splatting SLAM, using the detailed and varied Replica dataset. The findings demonstrate that while each algorithm exhibits unique strengths, Point-SLAM and SplaTAM generally outperform in terms of rendering and tracking accuracy. These algorithms provide robust solutions for precise 3D scene reconstruction, demonstrating their applicability across various real-world environments and challenges. Despite the thorough analysis presented, this study acknowledges several limitations. The computational efficiency and real-time processing capabilities of the algorithms were not exhaustively quantified, which are crucial factors for applications in robotics and augmented reality. The study also did not account for the variability in hardware performance, which can significantly influence the effectiveness of each algorithm. Future research should aim to address the noted limitations by expanding the evaluation framework to include outdoor environments and more dynamic scenes. By continuing to refine these algorithms, future studies can pave the way for broader applications and improvements in both existing and emerging fields.


\renewcommand{\bibfont}{\footnotesize}

\footnotesize{
\bibliographystyle{IEEEtran}
\bibliography{main}
}

\end{document}